\documentclass[10pt,twocolumn,letterpaper]{article}

\usepackage{iccv}
\usepackage{times}
\usepackage{epsfig}
\usepackage{graphicx}
\usepackage{amsmath}
\usepackage{amssymb}
\usepackage{graphicx}
\usepackage{amsmath}
\usepackage{amssymb}
\usepackage{booktabs}
\usepackage[accsupp]{axessibility}  % Improves PDF readability for those with disabilities.

\usepackage{url}
\usepackage[utf8]{inputenc}   % allow utf-8 input
\usepackage[T1]{fontenc}   % use 8-bit T1 fonts

\usepackage{url}   % simple URL typesetting
\usepackage{booktabs}   % professional-quality tables
\usepackage{amsfonts}   % blackboard math symbols
\usepackage{nicefrac}    % compact symbols for 1/2, etc.
\usepackage{microtype}   % microtypography
\usepackage{xcolor}   % colors
\usepackage{bbding}
\usepackage{pifont}
\usepackage{array}
\usepackage{wasysym}
\usepackage{amssymb}

\usepackage{url}
\usepackage[utf8]{inputenc} % allow utf-8 input
\usepackage[T1]{fontenc}    % use 8-bit T1 fonts
\usepackage{url}            % simple URL typesetting
\usepackage{booktabs}       % professional-quality tables
\usepackage{amsfonts}       % blackboard math symbols
\usepackage{nicefrac}       % compact symbols for 1/2, etc.
\usepackage{microtype}      % microtypography
\usepackage{xcolor}         % colors
\usepackage{tikz} 
\usepackage{pgfplots}
\usepackage{subcaption}

\usepackage{wrapfig} % allow utf-8 input
\usepackage{graphicx}

\usepackage{amsmath}
\usepackage{amssymb}
\usepackage{algorithm} %%算法
\usepackage{algorithmic} %%算法
\usepackage{enumerate} %%用于item
\usepackage{multirow} %%表格合并行单元格
\usepackage{amsmath}

\usepackage{helvet} %Required
\usepackage{courier} %Required
\usepackage{color}
\usepackage{dsfont}

\usepackage[textsize=tiny]{todonotes}

\newcommand{\Cmat}[0]{\ensuremath{{\bf C}} }

\newcommand{\Emat}[0]{\ensuremath{{\bf E}} }

\newcommand{\Lmat}[0]{\ensuremath{{\bf L}} }

\newcommand{\Pmat}[0]{\ensuremath{{\bf P}} }
\newcommand{\Qmat}[0]{\ensuremath{{\bf Q}} }

\newcommand{\Tmat}[0]{\ensuremath{{\bf T}} }

\newcommand{\ev}[0]{\ensuremath{\boldsymbol{e}} }

\newcommand{\lv}[0]{\ensuremath{\boldsymbol{l}} }

\newcommand{\ov}[0]{\ensuremath{\boldsymbol{o}} }
\newcommand{\pv}[0]{\ensuremath{\boldsymbol{p}} }

\newcommand{\xv}[0]{\ensuremath{\boldsymbol{x}} }
\newcommand{\yv}[0]{\ensuremath{\boldsymbol{y}} }

\newcommand{\Pimat}[0]{\ensuremath{\boldsymbol{\Pi}} }

\newcommand{\betav}[0]{\ensuremath{\boldsymbol{\beta}} }

\newcommand{\thetav}[0]{\ensuremath{\boldsymbol{\theta}} }

\usepackage[pagebackref=true,bookmarks=false]{hyperref}
\usepackage[capitalize,noabbrev]{cleveref}

\iccvfinalcopy % *** Uncomment this line for the final submission

\def\iccvPaperID{11734} % *** Enter the ICCV Paper ID here

\ificcvfinal\fi

\begin{document}

%%%%%%%%% TITLE
\title{PatchCT: Aligning Patch Set and Label Set with Conditional Transport\\ for Multi-Label Image Classification}

\author{Miaoge Li$^*$, Dongsheng Wang$^*$, Xinyang Liu, Zequn Zeng, Ruiying Lu, Bo Chen\\
National Key Laboratory of Radar Signal Processing\\
Xidian University, Xi’an, Shanxi 710071, China\\
{\tt\small {\{limiaoge, wds, xinyangatk\}@stu.xidian.edu.cn, bchen@mail.xidian.edu.cn}}
% For a paper whose authors are all at the same institution,
% omit the following lines up until the closing ``}''.
% Additional authors and addresses can be added with ``\and'',
% just like the second author.
% To save space, use either the email address or home page, not both
\and
Mingyuan Zhou\\
McCombs School of Business\\
The University of Texas at Austin, Austin, TX 78712, USA\\
{\tt\small mingyuan.zhou@mccombs.utexas.edu}
}
\maketitle
% Remove page # from the first page of camera-ready.
\ificcvfinal\thispagestyle{empty}\fi

\def\thefootnote{*}\footnotetext{Authors contributed equally.}
%%%%%%%%% ABSTRACT
\begin{abstract}
Multi-label image classification is a prediction task that aims to identify more than one label from a given image. This paper considers the semantic consistency of the latent space between the visual patch and linguistic label domains and introduces the conditional transport (CT) theory to bridge the acknowledged gap. While recent cross-modal attention-based studies have attempted to align such two representations and achieved impressive performance, they required carefully-designed alignment modules and extra complex operations in the attention computation. We find that by formulating the multi-label classification as a CT problem, we can exploit the interactions between the image and label efficiently by minimizing the bidirectional CT cost. Specifically, after feeding the images and textual labels into the modality-specific encoders, we view each image as a mixture of patch embeddings and a mixture of label embeddings, which capture the local region features and the class prototypes, respectively. CT is then employed to learn and align those two semantic sets by defining the forward and backward navigators. Importantly, the defined navigators in CT distance model the similarities between patches and labels, which provides an interpretable tool to visualize the learned prototypes. Extensive experiments on three public image benchmarks show that the proposed model consistently outperforms the previous methods. 
% Our code is available at \href{https://github.com/keepgoingjkg/PatchCT}{https://github.com/keepgoingjkg/PatchCT}.
\end{abstract}

%%%%%%%%% BODY TEXT
\section{Introduction}
\label{sec:intro}
Multi-label image classification is a fundamental yet challenging task in computer vision, where a set of objects needs to be predicted within one image. It has practical applications in wide fields such as image retrieval~\cite{wei2019saliency}, scene understanding~\cite{shao2015deeply}, recommendation~\cite{yang2015pinterest,jain2016extreme}, and biology analysis~\cite{ge2018chest,bai2022gaussian}. In addition to identifying the regions of interest, multi-label image classification also requires special attention on \textit{1)} semantic information of labels, \textit{e.g.}, label correlations, and \textit{2)} interactions between the visual image and textual label domains. Thus, it is often more complex and challenging compared to single-label case. 

\begin{figure}[!t]
\centering
\includegraphics[width=0.45\textwidth]{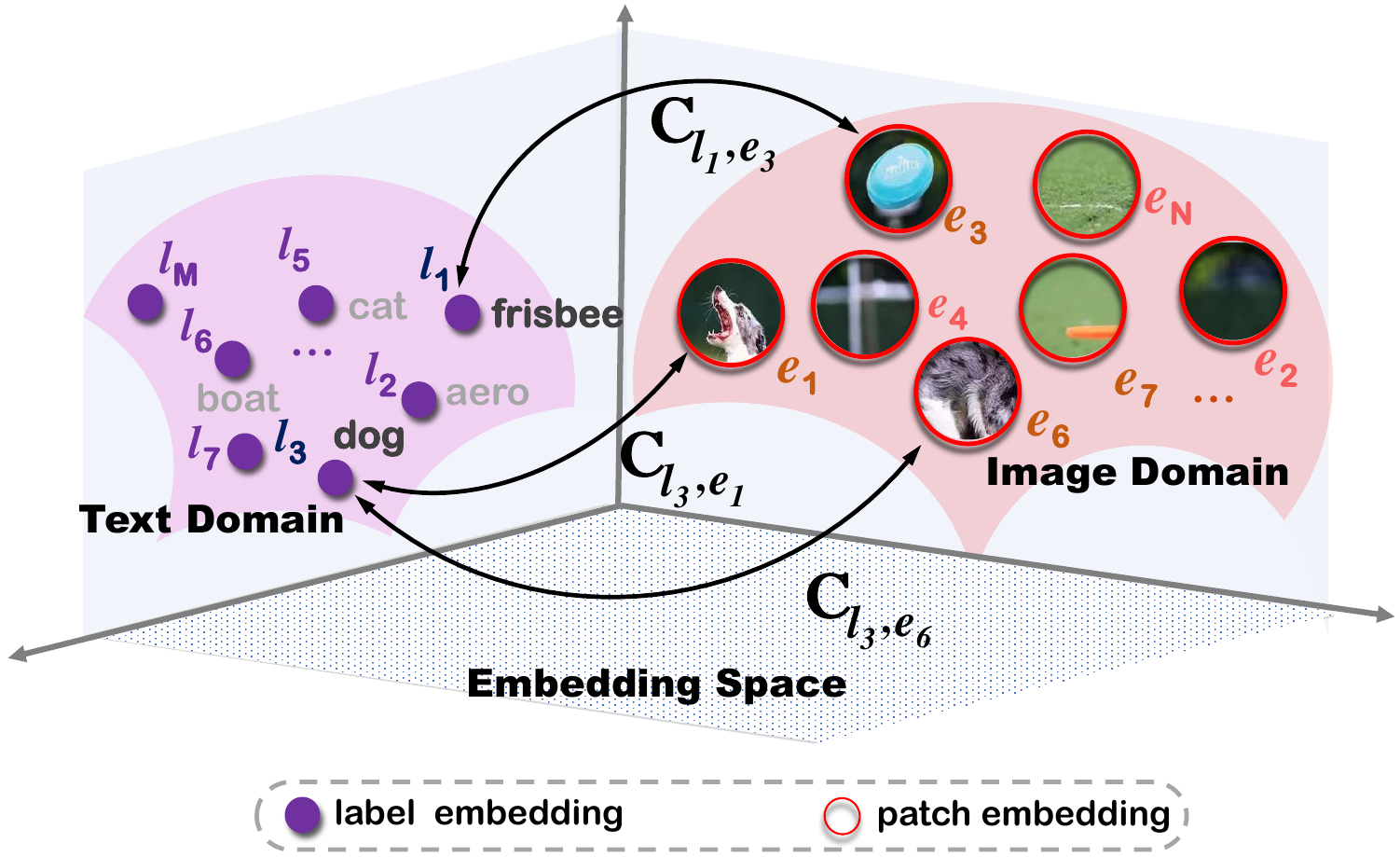}
\caption{\small{Motivation of the proposed PatchCT. We represent each image as a set of patch embeddings and a set of label embeddings, and then the conditional transport is employed as a semantic regularization to align such two domains.}}
\label{motivation}
\vspace{-5mm}
\end{figure}

To address the aforementioned challenges, several significant attempts have been developed from various directions. 
An increasing research attention is to learn semantic label representation. The core idea is intuitive: labels should be more correlated if they co-occurrence often. For instance, \textit{traffic lights} usually co-appeared with \textit{crosswalk}, and \textit{chair} has a high confidence score to appear if \textit{table} exists in the image.
% On one hand, acts as the prototype of that categories, a discriminative label embedding is beneficial for label prediction. On another hand, label dependency should be reflected in their embedding space, \textit{e.g.}, \textit{traffic lights} is usually co-appeared with \textit{crosswalk}, and \textit{chair} has a high confidence score to appear if \textit{table} exists in the image. 
Existing methods adopt pair-wise ranking loss, covariance matrices, recurrent structures, contrastive learning, graph neural networks (GCNs), and pre-trained language model to this end~\cite{zhang2013review, bi2014multilabel, wang2016cnn, yazici2020orderless, lanchantin2019neural, dao2021multi, bai2022gaussian, chen2019gcn, chen2019learning,jinye, liu2021query2label, zhu2022two}. Those methods either regular the learning with a pre-defined label graph which is often obtained from the training set or the pre-trained word embedding, \textit{e.g.} Glove~\cite{pennington2014glove}, or exploring label dependency implicitly (\textit{e.g.}, contrastive learning and BERT embeddings~\cite{bert}). Meanwhile, some works aim to solve the second issue by adopting cross-modal attention-based framework~\cite{huynh2020shared, liu2021query2label, lanchantin2021general, zhao2021m3tr,cheng2022mltr}, where a vision transformer (ViT~\cite{dosovitskiy2020image}) is often employed to obtain local patch features and the cross-attention between the labels and patches is then applied to align such two modalities. The alignment module acts as the core role in those models and needs to be designed carefully. Moreover, conventional attention mechanisms are guided by task-specific losses, without explicitly training signals to encourage alignment. The learned attention weights are therefore often dense and uninterpretable, leading to less effective relational prediction~\cite{chen2020graph}.

We address whether there is a more simple and principled approach to efficient alignments of vision-text domains. To explore this, in this paper, we introduce the conditional transport (CT) theory~\cite{zheng2021exploiting,tanwisuth2023pouf} and reformulate the multi-label classification task as a CT problem, where an image is represented as two discrete distributions over different supports, \textit{e.g.}, the visual patch domain and the textual label domain (as shown in Fig.~\ref{motivation}). 
Our idea is intuitive: those two distributions (or sets) are different modality representations of the same image, and therefore they share semantic consistency.
As a result, the learning of multi-label classification can be viewed as the process of aligning the textual label set to be as close to the visual set as possible. Accordingly, it is indeed key to find out how to measure the distance between two empirical distributions with different supports. Fortunately, conditional transport is well-studied in recent researches~\cite{zheng2021exploiting,tanwisuth2021prototype, tian2023prototypesoriented,wang2022representing} and provides a powerful tool for measuring the distance from one discrete distribution to another given the cost matrix. It is natural for us to develop a new framework for multi-label classification based on the minimization of CT distance.    

Specifically, the visual patch embeddings and textual label embeddings are first extracted by feeding the image and label descriptions into the corresponding encoders. We specify the image encoder as a ViT to capture spatial patch dependencies and the text encoder as a BERT to explore the label semantics.
% in order to extract the semantic correspondence between the vision and language modalities, we adopt the vision-language pre-trained (VLP) models, such as CLIP~\cite{clip} which is a self-supervised learning framework built on two modalities and shows great potential to promote image understanding guided by natural language supervision. The encoder of CLIP consists of two streams: the spatial stream that employs a ViT to capture spatial patch dependencies and a semantic stream for learning semantic label embeddings via a standard BERT. 
Besides, inspired by the current prompt learning success~\cite{liu2021gpt,coop}, we here design a simple and efficient template to reduce the domain shift between the language model and the multi-label classification and quickly distill the pre-trained knowledge to the downstream tasks simultaneously. After that, we collect patch features as a discrete probability distribution where each patch has a label-aware probability value that reflects the important score for multi-label prediction, \textit{e.g.} object patches have high probability values while background patches with lower attention. Similarly, we construct the textual label sets as a discrete distribution, whose probability values are obtained by normalizing the ground-truth label of the image. Given such two sets, the cost matrix in CT is then defined according to the similarity between the patches and labels, \textit{e.g.} the cosine distance of the patch and label embeddings. The CT divergence is defined with a bidirectional set-to-set transport, where a forward CT measures the transport cost from the patch set to the label set, and a backward CT that reverses the transport direction.
Therefore, by minimizing the bidirectional CT cost, we explicitly minimize the embedding distance between the domains, \textit{i.e.}, optimizing towards better patch-label alignments. Moreover, the learned transport plan models the semantic relations between patches and labels, which provides an interpretable tool to visualize the label concepts.
% The OT is then applied by a fast Sinkhorn algorithm~\cite{cuturi2013sinkhorn} to learn and align such two modalities. Thanks to the adopted CLIP, the cost matrix in OT can be defined by the similarity between the patch and label, \textit{e.g.} the inner product of the patch and label embeddings. Therefore, the learned transport plan achieves fine-grained matching across two sets and provides an interpretable tool to visualize the label embeddings. 

Our main contributions can be summarized as follows:
\begin{itemize}
    \item We propose a novel vision-text alignment framework based on conditional transport theory, where the interactions between the patches and labels are explored by minimizing the bidirectional CT distance between those two modalities to produce high semantic consistency for multi-label classification.
    \item We design sparse and layer-wise CT formulations to reduce the computational cost and enhance the deep interactions across modalities, contributing to robust and accurate alignments between patches and labels.
    \item Extensive experiments on three widely used visual benchmarks demonstrate the effectiveness of the proposed model by establishing consistent improvements on all data sets.
\end{itemize}

\section{Related Work}
\subsection{Multi-label Classification}
Multi-label classification has attracted increasing interest recently owing to its relevance in real-world applications. A natural idea comes from the single-label case and treats each category independently and then converts the task into a series of binary classification tasks straightforwardly. However, these methods often suffer from limited performance, due to their ignoring of the correlation between labels and objects’ spatial dependency which are quite crucial for multi-label image classification~\cite{wang2016cnn}. To address this issue, several previous proposed attempts explicitly capture the label dependencies by a CNN-based encoder followed by an RNN~\cite{wang2016cnn,chen2018order}. Apart from these sequential methods, some works resort to Graph Convolutional Networks(GCN) to model label relationships~\cite{chen2019learning, jinye}. However, it is also arguable that spurious correlations may be learned when the label statistics are insufficient. More recently, motivated by the great success of ViT in various visual tasks~\cite{dosovitskiy2020image,carion2020end,chen2021pre}, several works aim to align the labels and patches via the attention strategy for improving multi-label prediction. Those align-aware works are closest to our work. For example,
% Cheng \emph{et al}. \cite{cheng2022mltr} propose a Transformer architecture constructed with windows partitioning, in-window pixel attention, and cross-window attention, improving the performance particularly.
Query2Label of \cite{liu2021query2label} adopts the built-in cross-attention in the Transformer decoder as a spatial selector, in which label embeddings are treated as queries to align and pool class-related features from the ViT outputs for subsequent binary classifications. Lanchantin \emph{et al}.\cite{lanchantin2021general} utilize a general multi-label framework consisting of a Transformer encoder as well as a ternary encoding scheme during training to exploit the complex dependencies among visual features and labels. Zhao \emph{et al}.\cite{zhao2021m3tr} propose a multi-modal multi-label recognition transformer learning framework with three essential modules for complex alignments and correlations learning among inter- and intra-modalities. Different from those attention-based approaches that usually require carefully designed alignment modules and high computing costs in attention operations, we convert the multi-label classification task to a CT problem and align such two vision-label modalities by minimizing the total transport cost in a bidirectional view.

\subsection{Alignment via Transport Distance}
Recently, Optimal transport (OT)~\cite{villani2009optimal} has been used to solve the distance between two discrete distributions under unsupervised domain adaption~\cite{redko2019optimal}, label representation~\cite{frogner2015learning,zhao2018label,yang2018complex}, and cross-modal semantics~\cite{lee2019hierarchical, mahajan2019joint, chen2020graph}. For example, Lee \emph{et al}. introduce a hierarchical OT distance that leverages clustered structure in data to improve alignment between neural data and human movement patterns~\cite{lee2019hierarchical}. Chen \emph{et al}. formulate the cross-domain alignment as a graph matching problem and propose Graph OT (GOT)~\cite{chen2020graph} to various applications, including image caption, machine translation, and text summarization.
One of the core ideas behind those models is to align the multi-modalities by minimizing OT cost, they are distinct from ours in terms of task and technique detail. We focus on multi-label image classification cases where the local patch features and a set of class embeddings are considered under the CT framework. CT distance is originally developed to measure the difference between two probability distributions with the mini-batch optimization~\cite{zheng2021exploiting}. Unlike OT which usually needs to optimize the transport plan via an iterative Sinkhorn algorithm~\cite{cuturi2013sinkhorn}, CT considers the transport plan in a bidirectional view based on semantic similarity between samples from two distributions. This flexibility of CT potentially facilitates easier integration with deep neural networks with lower complexity and better scalability, showing better results on recent alignment tasks~\cite{tanwisuth2021prototype, wang2022representing,tanwisuth2023pouf,liu2023patch}. Those properties motivated us to learn aligned vision-label semantics under the CT framework for multi-label classification.

\begin{figure*}[!t]
\centering
\includegraphics[width=0.95\textwidth]{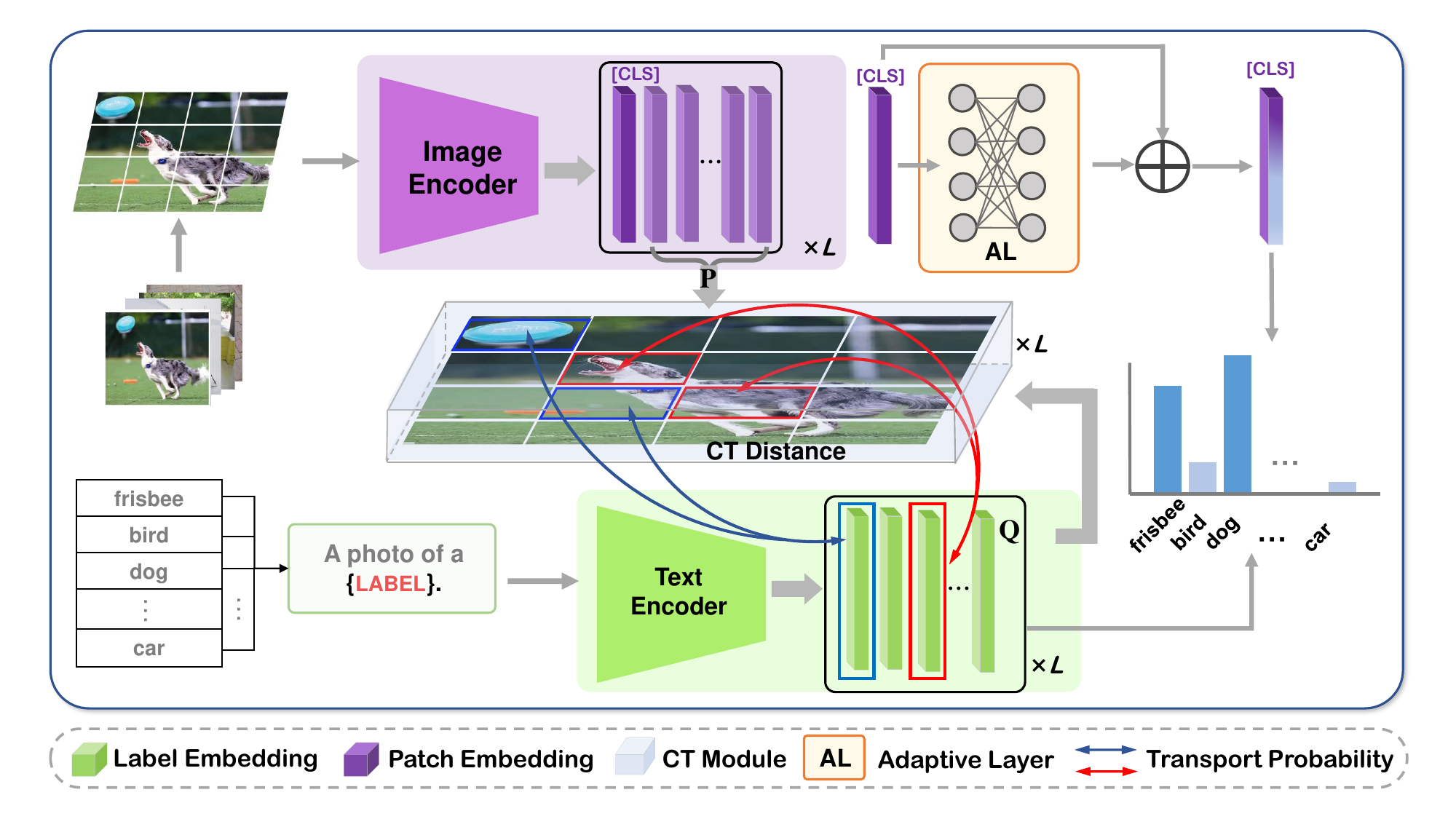}
\caption{\small{The overall framework of the proposed PatchCT, which loads the pre-trained ViT and BERT to capture visual patch and textual label embeddings. An adaptive module is added to transfer the knowledge implied in ViT to the multi-label classification task. A layer-wise CT distance is then applied to align the vision-text domains.}}
\label{framework}
\vspace{-3mm}
\end{figure*}

\section{Method}
In this section, we begin with the background and relevant notations of multi-label classification. Then we review the preliminaries of conditional transport and introduce the details of our proposed model, showing how to formulate the multi-label classification as a CT problem.

\subsection{Background and Notations}
Given a dataset with $I$ labeled samples together with a set of $M$ labels: $\mathcal{D}=\{ \xv_i, \yv_i \}_{i=1}^I$, where $\xv_i \in \mathbb{R}^{H \times W \times 3}$ denotes the $i$-th input RGB image with $H \times W$ size, and $\yv_i \in \{0, 1\}^{M}$ denotes the multi-hot binary label vector of $\xv_i$. $y_{mi} = 1$ means image $\xv_i$ have label $m$ and vice versa. Through observing $\mathcal{D}$, multi-label classification aims to derive a proper learning model, so that the label $\widetilde{\yv}$ of a test image $\xv$ can be predicted accordingly.

\subsection{Distance Between Two Set}
Let us consider two discrete probability distributions $\Pmat$ and $\Qmat \in \mathcal{P}(X)$ on space $X \in \mathbb{R}^{d}$: $\Pmat = \sum_{i=1}^n \theta_i \delta_{\ev_i}$, and $\Qmat = \sum_{j=1}^{m} \beta_j \delta_{\lv_j}$, where $\ev_i$ and $\lv_j$ are two points in the arbitrary same space $X$. $\thetav \in \Sigma^n$ and $\betav \in \Sigma^m$, the simplex of $\mathbb{R}^{n}$ and ~$\mathbb{R}^{m}$, denotes two probability values of the discrete states with the satisfy that $\sum_{i=1}^n \theta_i=1$ and $\sum_{j=1}^m \beta_j=1$. $\delta_{\ev}$ refers to a point mass located at coordinate $\ev \in \mathbb{R}^d$.

To measure such two discrete distributions, OT distance between $\Pmat$ and $\Qmat$ is formulated as the optimization problem:
% \begin{equation*}
$\text{OT}(\Pmat,\Qmat) = \min_{\Tmat \in \Pimat(\thetav, \betav)} \sum_{i,j} t_{ij}c_{ij}$, with $ \Tmat \mathds{1}^m=\thetav$, $\Tmat^T \mathds{1}^n = \betav.$
% \end{equation*}
Where $\mathds{1}^m$ is the $m$ dimensional vector of ones. $c_{ij} = c(\ev_i, \lv_j)\geq 0$ is the transport cost between the two points $\ev_i$ and  $\lv_j$ defined by an arbitrary cost function $c(\cdot)$. The optimal transport plan $\Tmat$ is often trained by minimizing the OT cost with the iterative Sinkhorn algorithm~\cite{cuturi2013sinkhorn}.
% As suggested in \citep{cuturi2013sinkhorn}, an entropic constraint is introduced into the above distance: $H = - \sum_{ij} t_{ij} \ln t_{ij}$,
% resulting in the iterative Sinkhorn algorithm to reduce the time complexity. 

More recently, the demand on efficient computation and bidirectional asymmetric transport promotes the development of CT divergence \cite{zheng2021exploiting}, which can be applied to quantify the difference between discrete empirical distributions in various applications \cite{zheng2021contrastive, tanwisuth2021prototype, wang2022representing}. Specifically, given the above source and target distributions $\Pmat$ and $\Qmat$, the CT cost is defined with a bidirectional distribution-to-distribution transport, where a forward CT measures the transport cost from the source to target, and a backward CT that reverses the transport direction. Therefore, the CT problem can be defined as:
\begin{equation} \label{ct}
    \text{CT}_{\Cmat}(\Pmat, \Qmat)= \min _{ \overrightarrow{\Tmat} \overleftarrow{\Tmat}}  (\sum_{i,j} \overrightarrow{t}_{ij} c_{ij} + \sum_{j,i} \overleftarrow{t}_{ji} c_{ji}),
\end{equation}
where $\Cmat$ is the cost matrix, and $\overrightarrow{t}_{ij}$ in $\overrightarrow{\Tmat}$ acts as the transport probability (the navigator) from the source point $\ev_i$ to the target point $\lv_j$: $\overrightarrow{t}_{ij} = \theta_i \frac{\beta_j e^{-d_{\psi}(\ev_{i},\lv_{j})}}{\sum_{j^{\prime}=1}^{m} \beta_{j^{\prime}} e^{-d_{\psi}(\ev_i, \lv_{j^{\prime}})}}$, hence $ \overrightarrow{\Tmat} \mathds{1}^{m}=\thetav$. Similarly, we have the reversed transport probability: $\overleftarrow{t}_{ji} = \beta_j \frac{\theta_i e^{-d_{\psi}(\lv_{j} , \ev_{i})}}{\sum_{i^{\prime}=1}^{n} \theta_{i^{\prime}} e^{-d_{\psi}(\lv_j, \ev_{i^{\prime}})}}$, and $\overleftarrow{\Tmat} \mathds{1}^n=\betav$. 
%Note that 
The distance function $d_\psi(\ev_i, \lv_j)$ parameterized with $\psi$ can be implemented by deep neural networks to measure the semantic similarity between two points, making CT amenable to 
%mini-batch 
stochastic gradient descent-based optimization.

\subsection{The Proposed Framework}
Now we present the details of our proposed PatchCT, which aligns visual \textbf{Patch} and textual label domains under \textbf{CT} framework for multi-label image classification. As shown in Fig.~\ref{framework}, PatchCT consists of four components, the visual $\Pmat$ set, the textual label $\Qmat$ set, the adaptive layer, and the CT distance between $\Pmat$ and $\Qmat$.  

\paragraph{$\Pmat$ set over patch embeddings.} For an input image $\xv_i$, PatchCT first divides it into $N$ patches evenly and then feeds them to the image encoder to obtain the local region features $\{ \ev_n \}_{n=1}^N$ (we omit $i$ for convenience), where $\ev_n \in \mathbb{R}^{d}$ denotes the $n$-th patch embedding, $d$ is the embedding dimension. In addition to the $N$ local features, PatchCT also learns the [$\textit{CLS}$] visual token $\ev_{\textit{CLS}}$ that acts as the global image representation. One of the main operations in the image encoder is the multi-layer multi-head attention layers that integrate and update the global and local features by considering the spatial and contextual information among patches. 
Traditional CT settings often view each point equally, $\textit{e.g.}$, $\thetav_i$ is a uniform distribution over $N$ points. Unfortunately, this is not the case in multi-label tasks where only a few key patches contribute to the final prediction practically. To this end, PatchCT defines a sparse and label-guided $\thetav_i$ as below:
\begin{equation} \label{theta}
    \thetav_i = \text{softmax}(\text{TopK}(\Emat_i^T \ov_i, k)), \quad \ov_i = \Lmat \hat{\yv}_i   
\end{equation}
where $\Emat_i \in \mathbb{R}^{d \times N}$ and $\Lmat \in \mathbb{R}^{d \times M}$ are the patch embedding matrix of $\xv_i$ and label embedding matrix respectively. $\hat{\yv}_i$ is the normalized label vector $\hat{\yv_i} = \yv_i / \sum_{m=1}^M y_{mi}$, thus $\ov_i$ here denotes the label-aware representation of $\xv_i$, which is used to select core patches that have close semantics with the ground-truth labels. $\text{TopK}(\cdot, k)$ is a sparsity operation that masks top-$k$  patches with 1 and 0 for others based on the similarity score, and $\text{softmax}(\cdot)$ makes sure the probability simplex of $\thetav_i$.

After giving the patch embedding matrix $\Emat_i$ and its weights $\thetav_i$, PatchCT obtains the discrete distribution $\Pmat_i$ of $\xv_i$ over the visual set. Note that $\Pmat_i$ collects the detailed visual features of local patches, thus it brings benefits to the downstream multi-label task. For one thing, $\thetav_i$ in Eq.~\ref{theta} guarantees the sparsity of $\Pmat_i$ which is useful for reducing the computing cost and enhancing the interpretability of PatchCT. For another, $\Pmat_i$ concentrates more on the core patches that contain objects via the introduced label-aware selection strategy, leading to more discriminative features.

\paragraph{$\Qmat$ set over textual label embeddings.} In addition to the visual set $\Pmat$, PatchCT also represents each image as a set of textual label embeddings. Inspired by the recent success of prompt learning, we perform a simple but efficient prompt: \textit{A photo of \{label\} .} on each label and acquire $M$ label sentences. One can obtain the label embeddings $\Lmat = \{ \lv_m | _{m=1}^M \} \in \mathbb{R}^{d \times M}$. Due to the accessible label vector during training, it is natural to define $\betav$ in $\Qmat$ by normalizing $\yv_i$:
\begin{equation*}
    \betav_i = softmax(\yv_i).
\end{equation*}
This means that $\Qmat_i$ collects the ground-truth label features as the textual representation of $\xv_i$. Besides, with the semantic linguistic knowledge implied in the pre-trained text encoder (\textit{e.g., BERT}) and the designed prompt templates, the learned label embeddings have the ability to capture \textit{1)} the textual semantics for each label, \textit{2)} the correlations among labels, which help to improve the identification of label representations~\cite{liu2021query2label}.

To summarize briefly, the proposed PatchCT views the multi-label classification as a CT problem and formulates each image $\xv_i$ with two discrete distributions $\Pmat_i$ and $\Qmat_i$ over the visual patch and textual label embeddings respectively. Those two representations share semantic consistency but with different supports. One of the core ideas behind PatchCT is to align the vision-text modalities by minimizing the bidirectional CT distance of $\Pmat$ and $\Qmat$ for multi-label prediction, which will be discussed in Sec.~\ref{loss}.

\paragraph{Adaptive layer of visual features.} Like previous works~\cite{zhao2021m3tr}, we here adopt the pretrained ViT as our image encoder to obtain the local patch embeddings. To further enhance the representation of local patches and distill the pre-trained knowledge of ViT in an efficient way for multi-label classification, PatchCT introduces an adaptive module $f_\phi$ following the last layer of the image encoder~\cite{clip_ada}, where $\phi$ is the learnable parameters in $f_{\phi}$. Note that $f_\phi$ is a lightweight network that consists of two linear layers and aims to adapt ViT features to \textit{new} knowledge that is more suitable for the multi-label task. Moreover, we also apply the residual connection of the adapted features and the original features encoded by the pre-trained ViT for mixing the two pieces of information. Thus the global image feature can be updated as (we still use $\xv_i$ for convenience):
\begin{equation} \label{ada_x}
    \xv_i = \ev_{\text{CLS}} \oplus f_\phi (\ev_{\text{CLS}}),
\end{equation}
where $\oplus$ denotes the residual connection. $\xv_i$ acts as the visual feature and is used to make the multi-label prediction. 

\subsection{The objective function} \label{loss}
The final objective function to minimize is simply the summation of the layer-wise CT distance that aligns the vision-text domains and the asymmetric loss for multi-label classification.
\paragraph{Layer-Wise CT Distance.}
For image $\xv_i$, the two discrete distributions $\Pmat_i, \Qmat_i$ are semantic representations from two different domains. PatchCT bridges the semantic gaps by minimizing the CT distance of $\Pmat_i$ and $\Qmat_i$, \textit{e.g.}, $\text{CT}(\Pmat_i, \Qmat_i)$. To make deep interactions between such two modalities and achieve better alignments, we develop a Layer-wise CT distance (LCT) that greedily minimizes the CT distance at each layer. With a $L$-layer domain-specific encoders, the LCT can be expressed as:
\begin{equation*}
    \text{LCT}^{(L)} = \sum_{l=1}^L \text{CT}(\Pmat_i^{(l)}, \Qmat_i),
\end{equation*}
where $l$ is the index of layer, $\thetav_i^{(l)}$ is calculated according to Eq.~\ref{theta} by replacing the patch embedding matrix and label-aware embedding at the corresponding layer $l$. The cost matrix $\Cmat^{(l)}$ in Eq.~\ref{ct} is computed by the cosine distance:
\begin{equation*}
    \Cmat_i^{(l)} = 1 - \frac{ {\Emat_i^{(l)}}^T \Lmat^{(l)} }{ || \Emat_i^{(l)} || || \Lmat^{(l)} || } 
\end{equation*}

\paragraph{Asymmetric loss.}
Given the mixed visual embedding $\xv_i$ calculated by Eq.~\ref{ada_x} and the aligned label embedding matrix at $L$-th layer $\Lmat^{(L)}$, one can predict the category probabilities of $i$-th image as $\pv_i = \sigma ( {\Lmat^{(L)}}^T \xv_i) \in \mathbb{R}^M$, where $\sigma(\cdot)$ is the sigmoid function. To more effectively address the label imbalance issue, we adopt the asymmetric loss (ASL), which is a variant of focal loss and a valuable choice in multi-label classification tasks~\cite{asl, liu2021query2label}:
\begin{equation*}
    \text{ASL} = \frac{1}{M} \sum_{m=1}^M \left\{
    \begin{array}{ll}
         (1-p_{mi})^{\gamma +} \text{log}(p_{mi}), \quad &y_{mi}=1, \\
         (p_{mi})^{\gamma -} \text{log}(1-p_{mi}), \quad &y_{mi} = 0,
    \end{array} \right.
\end{equation*}
where $\gamma+, \gamma-$ are two hyper-parameters for positive and negative values.

Let the learnable parameters as $\Omega=\{\text{Enc}, \phi, \psi\}$ that denotes the parameters in image and text encoders, the adaptive module, the defined transport plan in Eq.~\ref{ct}, respectively. $\Omega$ is optimized using stochastic gradient descent by minimizing the combined loss:
\begin{equation}
    \mathcal{L} = \text{LCT} + \text{ASL},
\end{equation}
where the first term ensures the alignments between the vision and text modalities at each layer in two domain-specific encoders and the second term provides supervised information for multi-label classification. %\rr{We summarize the training algorithm at Alg.~\ref{alg}.}

\begin{table*}[!t]
    \centering
    \scalebox{0.88}{
    \begin{tabular}{c|c|c|c|c|c|c|c|c|c|c|c|c|c}
    \toprule
    \multirow{2}*{Methods} & \multicolumn{6}{c|}{ALL} &\multicolumn{6}{c|}{Top-3} & \multirow{2}*{mAP}\\
    \cline{2-13}
         &CP &CR &CF1 &OP &OR &OF1 &CP &CR &CF1 &OP &OR &OF1  \\
         \hline
         \hline
        CNN-RNN~\cite{wang2016cnn} &- &- &- &- &- &- &66.0 &55.6 &60.4 &69.2 &66.4 &67.8 &61.2\\
        
        ResNet-101~\cite{DBLP:conf/cvpr/HeZRS16} &80.2 &66.7 &72.8 &83.9 &70.8 &76.8 &84.1 &59.4 &69.7 &89.1 &62.8 &73.6 &77.3\\
        
        ML-GCN~\cite{chen2019multi} &85.1 &72.0 &78.0 &85.8 &75.4 &80.3 &89.2 &64.1 &74.6 &90.5 &66.5 &76.7 &83.0\\
        
        SSGRL†~\cite{chen2019learning} &\textbf{89.5} &68.3 &76.9 &\textbf{91.2} &70.7 &79.3 &91.9 &62.1 &73.0 &\textbf{93.6} &64.2 &76.0 &83.6\\
        
        CMA~\cite{you2020cross} &82.1 &73.1 &77.3 &83.7 &76.3 &79.9 &87.2 &64.6 &74.2 &89.1 &66.7 &76.3 &83.4\\
        
        TSGCN~\cite{xu2020joint} &81.5 &72.3 &76.7 &84.9 &75.3 &79.8 &84.1 &67.1 &74.6 &89.5 &69.3 &69.3 &83.5\\
        
        MulCon~\cite{dinhcontrast}&- &- &78.6 &- &- &81.0 &- &- &- &- &- &- &84.0\\
        
        C-Tran†~\cite{lanchantin2021general} &86.3 &74.3 &79.9 &87.7 &76.5 &81.7 &90.1 &65.7 &76.0 &92.1 &\textbf{71.4} &77.6 &85.1\\
        
        ADD-GCN~\cite{ye2020attention} &84.7 &75.9 &80.1 &84.9 &79.4 &82.0 &88.8 &66.2 &75.8 &90.3 &68.5 &77.9 &85.2\\
        
        ASL~\cite{ridnik2021asymmetric} &87.2 &76.4 &81.4 &88.2 &79.2 &81.8 &91.8 &63.4 &75.1 &92.9 &66.4 &77.4 &86.6\\
       
        CSRA~\cite{zhu2021residual} &89.1 &74.2 &81.0 &89.6 &77.1 &82.9 &\textbf{92.5} &65.8 &76.9 &93.4 &68.1 &78.8 &86.9\\
        
        Q2L~\cite{liu2021query2label} &87.6 &76.5 &81.6 &88.4 &78.5 &83.1 &91.9 &66.2 &77.0 &93.5 &67.6 &78.5 &87.3\\
        
        M3TR~\cite{zhao2021m3tr} &88.4 &77.2 &82.5 &88.3 &79.8 &83.8 &91.9 &68.1 &78.2 &92.6 &69.6 &79.4 &87.5\\
        % TSFormer~\cite{zhu2022two} &88.3 &79.2 &83.5 &88.5 &81.5 &84.9 &92.3 &69.1 &79.0 &93.2 &70.5 &80.3 &88.9\\
        \hline
        PatchCT &83.3 &\textbf{82.3} &\textbf{82.6} &84.2 &\textbf{83.7} &\textbf{83.8} &90.7 &\textbf{69.7} &\textbf{78.8} &90.3 &70.8 &\textbf{79.8} &\textbf{88.3} \\
        \bottomrule
    \end{tabular}}
    \caption{\small{Comparison of PatchCT and known SOTA models on MS-COCO dataset under the settings of all and top-3 labels. All metrics all in \%. The symbol † means using a larger input image resolution (576 × 576).}}
    \label{coco}
\end{table*}

\section{Experiments}
In this section, we evaluate our PatchCT with known state-of-the-art methods on three widely-used multi-label image benchmarks by reporting a series of metrics. In addition to the numerical results, comprehensive ablation and qualitative studies of the proposed model are also provided. Our code is available at \href{https://github.com/keepgoingjkg/PatchCT}{https://github.com/keepgoingjkg/PatchCT}.
\subsection{Datasets and Evaluation Metrics}
\paragraph{Datasets.} We conduct extensive experiments on three popular image datasets, including \textbf{MS-COCO}~\cite{lin2014microsoft}, \textbf{PASCAL VOC 2007}~\cite{everingham2010pascal}, and \textbf{NUS-WIDE}~\cite{chua2009nus}. MS-COCO is a  commonly-used benchmark to evaluate the multi-label image classification task. It contains 82,081 images as the training set and 40,504 images as the validation set. The objects are categorized into 80 classes with about 2.9 object labels per image. Pascal VOC 2007 contains 20 categories and total 9,963 images, in which 5,011 images form the train-val set, and the remaining 4,952 images are taken as the test set for evaluation. For fair comparisons, the current competitors and our model are all trained on the train-val set and evaluated on the test set. NUS-WIDE is a real-world web image dataset with 269,648 images and 5018 labels from Flickr. These images are further manually annotated with 81 visual concepts. After removing the no-annotated images, the training set and test set contain 125,449 and 83,898 images respectively. We summarize the statistics of datasets at Table.~\ref{dataset}. 

\begin{table}[!t]
    \centering
    \scalebox{0.9}{
    \begin{tabular}{c|c|c|c|c}
        \toprule[1.5pt]
         Datasets & \# Train &\# Test &\# Class &\# Object\\
         \hline 
         MS-COCO &82,081 &40,504 & 80 & 2.9\\
         \hline
         VOC 2007 &5,011 &4952 &20 &2.5 \\
         \hline
         NUS-WIDE &125,449 &83898 &81 &2.4 \\
         \bottomrule
    \end{tabular}}
    \caption{\small{Statistics of the used datasets. \# Object denotes the average object labels per image.}}
    \label{dataset}
\end{table}

\paragraph{Evaluation Metrics.} To comprehensively evaluate the performance, we follow previous works~\cite{wang2016cnn, chen2019gcn, liu2021query2label} and report the mean average precision(mAP), the average per-class precision (CP), recall (CR), F1 (CF1), and the average overall precision(OP), recall(OR), F1(OF1). Besides, we also report the results of top-3 labels. For all metrics, higher values indicate better performance, and in general, mAP is the most important metric. For each image, labels are identified as positive if their predicted probabilities are greater than 0.5. All metrics are reported as the mean results of five runs with different random seeds.

\begin{table*}[!t]
    \centering
    \setlength\tabcolsep{2pt} 
    \scalebox{0.85}{
    \begin{tabular}{c|cccccccccccccccccccc|c}
        \toprule
         Methods &aero &bike &bird &boat &bottle &bus &car &cat &chair &cow &table &dog &horse &motor &person &plant &sheep &sofa &train &tv &mAP \\
         \hline
         \hline
         CNN-RNN~\cite{wang2016cnn} &96.7 &83.1 &94.2 &92.8 &61.2 &82.1 &89.1 &94.2 &64.2 &83.6 &70.0 &92.4 &91.7 &84.2 &93.7 &59.8 &93.2 &75.3 &99.7 &78.6 &84.0\\

         RLSD~\cite{zhang2018multilabel} &96.4 &92.7 &93.8 &94.1 &71.2 &92.5 &94.2 &95.7 &74.3 &90.0 &74.2 &95.4 &96.2 &92.1 &97.9 &66.9 &93.5 &73.7 &97.5 &87.6 &88.5\\
         HCP~\cite{wei2015hcp} &98.6 &97.1 &98.0 &95.6 &75.3 &94.7 &95.8 &97.3 &73.1 &90.2 &80.0 &97.3 &96.1 &94.9 &96.3 &78.3 &94.7 &76.2 &97.9 &91.5 &90.9\\

         RDAR~\cite{wang2017multi} &98.6 &97.4 &96.3 &96.2 &75.2 &92.4 &96.5 &97.1 &76.5 &92.0 &87.7 &96.8 &97.5 &93.8 &98.5 &81.6 &93.7 &82.8 &98.6 &89.3 &91.9\\
         RARL~\cite{chen2018recurrent} &98.6 &97.1 &97.1 &95.5 &75.6 &92.8 &96.8 &97.3 &78.3 &92.2 &87.6 &96.9 &96.5 &93.6 &98.5 &81.6 &93.1 &83.2 &98.5 &89.3 &92.0\\

         SSGRL†~\cite{chen2019learning} &99.5 &97.1 &97.6 &97.8 &82.6 &94.8 &96.7 &98.1 &78.0 &97.0 &85.6 &97.8 &98.3 &96.4 &98.8 &84.9 &96.5 &79.8 &98.4 &92.8 &93.4\\
         ML-GCN~\cite{chen2019multi} &99.5 &98.5 &98.6 &98.1 &80.8 &94.6 &97.2 &98.2 &82.3 &95.7 &86.4 &98.2 &98.4 &96.7 &99.0 &84.7 &96.7 &84.3 &98.9 &93.7 &94.0\\
         TSGCN~\cite{xu2020joint} &98.9 &98.5 &96.8 &97.3 &87.5 &94.2 &97.4 &97.7 &84.1 &92.6 &89.3 &98.4 &98.0 &96.1 &98.7 &84.9 &96.6 &87.2 &98.4 &93.7 &94.3\\
         ASL~\cite{ridnik2021asymmetric} &- &- &- &- &- &- &- &- &- &- &- &- &- &- &- &- &- &- &- &- &94.6\\
         CSRA~\cite{zhu2021residual} &99.9 &98.4 &98.1 &98.9 &82.2 &95.3 &97.8 &97.9 &84.6 &94.8 &90.8 &98.1 &97.6 &96.2 &99.1 &86.4 &95.9 &88.3 &98.9 &94.4 &94.7\\
         MlTr-1~\cite{cheng2022mltr} &- &- &- &- &- &- &- &- &- &- &- &- &- &- &- &- &- &- &- &- &95.8\\
         Q2L~\cite{liu2021query2label} &99.9 &98.9 &99.0 &98.4 &\textbf{87.7} &98.6 &98.8 &99.1 &84.5 &98.3 &89.2
         &99.2 &99.2 &\textbf{99.2} &\textbf{99.3} &90.2 &98.8 &88.3 &99.5 &95.5 &96.1 \\
         M3TR~\cite{zhao2021m3tr} &99.9 &99.3 &\textbf{99.1} &99.1 &84.0 &97.6 &98.0 &99.0 &85.9 &\textbf{99.4} &93.9 &\textbf{99.5} &99.4 &98.5 &99.2 &90.3 &\textbf{99.7} &91.6 &\textbf{99.8} &96.0 &96.5\\
         % TSFormer~\cite{zhu2022two} &100.0 &99.2 &99.2 &98.6 &86.4 &97.2 &98.4 &98.9 &88.9 &99.5 &95.3 &99.7 &99.6 &99.1 &99.4 &90.0 &99.6 &93.7 &99.9 &96.7 &97.0\\
         \hline
         PatchCT &\textbf{100.0} &\textbf{99.4} &98.8 &\textbf{99.3} &87.2 &\textbf{98.6} &\textbf{98.8} &\textbf{99.2} &\textbf{87.2} &99.0 &\textbf{95.5} &99.4 &\textbf{99.7} &98.9 &99.1 &\textbf{91.8} & 99.5 &\textbf{94.5} &99.5 &\textbf{96.3} &\textbf{97.1}\\
         \bottomrule
    \end{tabular}}
    \caption{Results on Pascal VOC 2007 dataset in terms of class-wise precision (AP) and mean average precision (mAP) in \%.}
    \label{voc}
\end{table*}

\begin{table}[!t]
    \centering
    \begin{tabular}{c|c|c|c|c|c}
        \toprule
         \multirow{2}*{Methods} &\multicolumn{2}{c|}{ALL} &\multicolumn{2}{c|}{Top-3} & \multirow{2}*{mAP}  \\
         \cline{2-5}
         &CF1 &OF1 &CF1 &OF1 \\
         \hline
         \hline
         CNN-RNN~\cite{wang2016cnn} &- &- &34.7 &55.2 &56.1\\
         \hline ResNet-101~\cite{DBLP:conf/cvpr/HeZRS16} &51.9 &69.5 &56.8 &69.1 &59.8\\
        \hline     CMA~\cite{you2020cross} &60.5 &73.7 &55.5 &70.0 &61.4\\
         \hline
        MlTr-1~\cite{cheng2022mltr} &65.0 &\textbf{75.8} &- &- &66.3\\
        \hline
        MulCon~\cite{dinhcontrast} &59.0 &73.8 &- &- &62.5\\
        \hline
        
        ASL~\cite{ridnik2021asymmetric} &63.6 &75.0 &- &- &65.2\\
        \hline
        Q2L~\cite{liu2021query2label} &64.0 &75.0 &- &- &66.3\\
         % TSFormer~\cite{zhu2022two} &64.9 &76.0 &59.6 &70.7 &69.3 \\
         \hline
         PatchCT &\textbf{65.5} &74.7 &\textbf{61.2} &\textbf{71.0} &\textbf{68.1}\\
         \bottomrule
    \end{tabular}
    \caption{Results on the NUS-WIDE dataset under the setting of all and top-3 labels. All metrics in \%.}
    \label{nus-wide}
\end{table}

\subsection{Implementation Details}
Like previous works, PatchCT employs the ViT-B16 that pretrained on ImageNet21k as the image encoder and loads the 12-layer BERT trained on Wikipedia data as the text encoder, and the hidden dimension in both encoders is $d=512$. For a fair comparison with competitors, we resize all images to $ H \times W = 448 \times 448$ as input resolution in both training and testing phases throughout all experiments. The number of top-$k$ patches is set as $k=200$. We set $\gamma+=0$ and $\gamma-=2$ in the asymmetric loss. The optimization of PatchCT is done by AdamW with a learning rate of $1e-5$, True-weight-decay $1e-2$, and batch size 12 for maximally 40 epochs. All experiments are performed on one Nvidia RTX-3090Ti GPU and our model is implemented in PyTorch. Please refer to the attached code for more details.

\subsection{Comparison with State-of-the-art Methods}
In order to demonstrate the effectiveness of the proposed framework, we compare PatchCT with a number of state-of-the-art methods from the literature, including conventional CNN-based methods~\cite{wang2016cnn, DBLP:conf/cvpr/HeZRS16,zhang2018multilabel}, graph-based methods~\cite{chen2019multi, chen2019learning,xu2020joint}, and attention-based methods~\cite{zhu2021residual,cheng2022mltr,lanchantin2021general,liu2021query2label,zhao2021m3tr}. The numbers of the compared methods are taken from the best-reported results in their papers (We report results of Q2L with TResNet-L to keep close compute and numbers of parameters fairly with the image encoder of PatchCT).
Table.(~\ref{coco}-\ref{nus-wide}) report the comparisons of PatchCT with those SOTA methods on the three multi-label image datasets. We have the following remarks about the numerical results: 
\textit{1)} Overall, our proposed PatchCT achieves the best mAP scores on all datasets. Despite scoring high on some metrics, other approaches often either fail to balance accuracy and recall or fail to achieve stable performance on all categories (\textit{e.g.}, chair, table, horse). 
\textit{2)} Compared to models that leverage additional information of label dependency~\cite{chen2019gcn, chen2019multi, xu2020joint}, PatchCT accomplishes significant improvements in most metrics. This suggests the efficiency of the employed pre-trained language model that contains rich semantic knowledge and shows great potential to capture label correlations. Besides, the AP scores of our PatchCT on all 20 categories in Pascal VOC 2007 dataset are above 87.2, and PatchCT exceeds 97.1 mAP on the test set. This demonstrates that the text encoder can also provide discriminative label representations.
\textit{3)} Furthermore, developed from the similar motivation that aligns the vision-text modalities for the multi-label task, our PatchCT shows a relatively stable and superior performance compared to recent attention-based models~\cite{liu2021query2label,zhao2021m3tr, lanchantin2021general,cheng2022mltr} that adopt the cross-attention mechanism to explore the interactions between those two domains. We attribute this success to the sparsity layer-wise CT distance. Further, LCT provides an efficient option to align the semantics of patches and labels progressively by minimizing the transport cost from a bidirectional view. Moreover, we also visualize the learned transport probabilities at Sec.~\ref{vis} which indicates the interpretability of PatchCT.

\begin{figure*}[!t]
\centering
\includegraphics[width=0.95\textwidth]{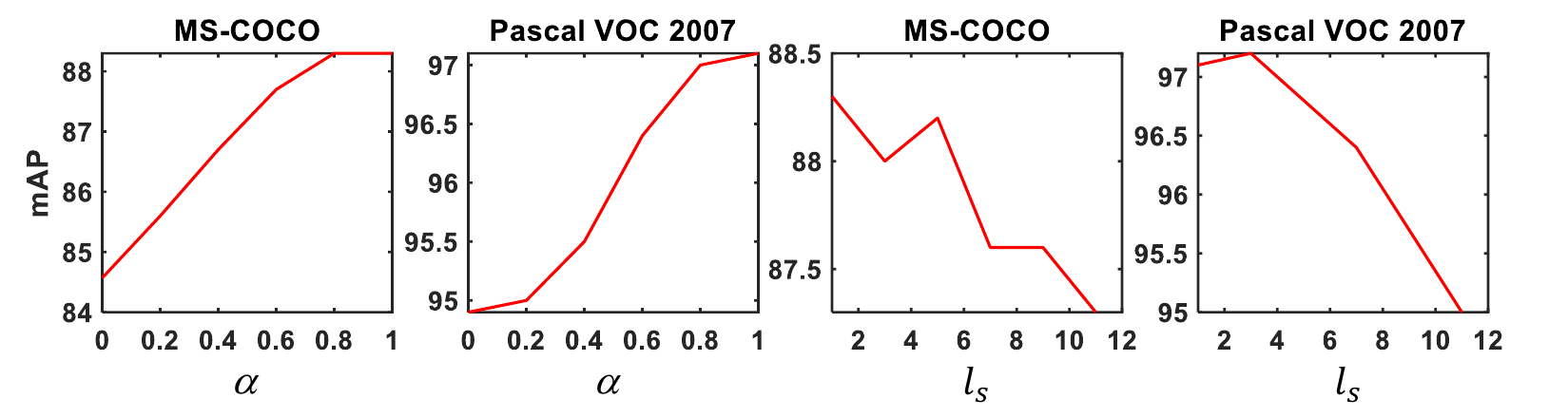}
\vspace{-2mm}
\caption{\small{Parameter sensitivity of PatchCT in terms of $\alpha$} that controls the weights of LCT in the combined loss (the first two subfigures) and the starting layer of the CT alignments $l_s$ (the last two subfigures) on MS-COCO and Pascal VOC 2007.}
\label{abla}
% \vspace{-5mm}
\end{figure*}

\begin{figure*}[!t]
\centering
\includegraphics[width=0.95\textwidth]{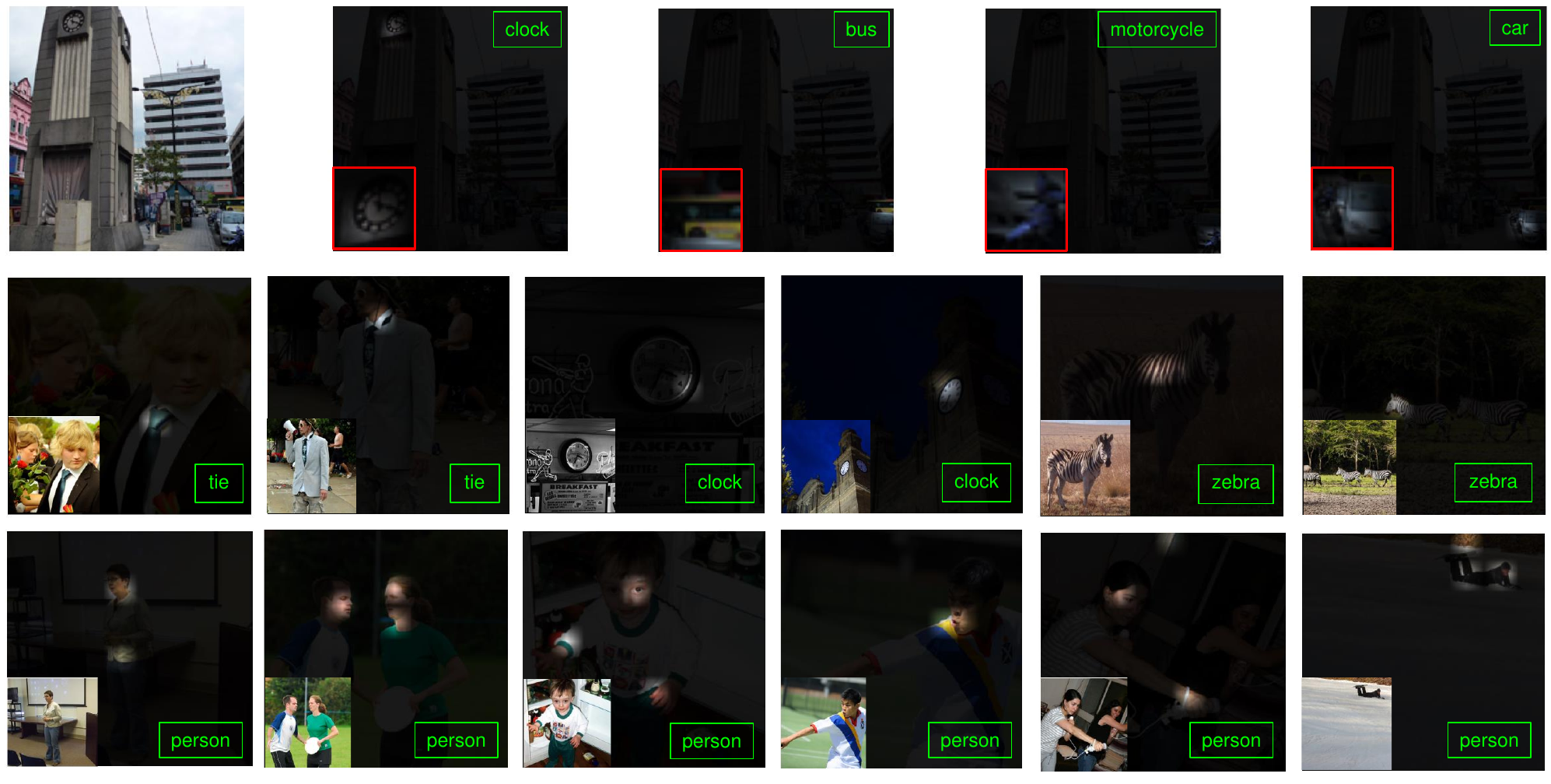}
\caption{\small{Visualization of the learned backward transport probabilities. The top row includes the input image and the different label queries. The activated areas are zoomed in for clear display (red boxes). The bottom two rows show activated patches in different images for labels ``tie'', ``clock'', ``zebra'', and ``person''. The raw images are included in the bottom left corner.} }
\label{atten}
\vspace{-3mm}
\end{figure*}

\subsection{Further Analyses}\label{ablas}
To fully understand the introduced CT module in PatchCT, we perform a series of further analyses in this section, Specifically, we are interested in the following variants.

\paragraph{Effects of the CT loss.} We treat the LCT distance and the asymmetric loss equally in previous experiments for convenience. Here we adopt a weighted combined loss: $\mathcal{L}_{\alpha} = \alpha \text{LCT} + \text{ASL}$, where $\alpha$ controls the weight of the LCT. We report the mAP of PatchCT under the setting of $\alpha=[0:0.2:1]$ on MS-COCO and VOC 2007 datasets at the first two subfigures of Fig.~\ref{abla}. Note that, $\alpha=0$ means we remove LCT from the combined loss, which results in our base model without CT alignment. The results of this variant are also reported at Table~\ref{rebuttal} for a clear comparison. We find that The introduced LCT plays a pivotal role in PatchCT and improves the performance by a significant margin, which demonstrates the efficiency of the vision-text alignment by minimizing the bidirectional transport cost.

\paragraph{Effects of the starting layer in LCT.}
Due to its irreplaceable role in PatchCT, we further explore the effects of the starting layer $l_s$ in LCT and report the mAP scores on MS-COCO and VOC 2007 with the setting $l_s=[1:2:11]$ at the last two subfigures of Fig.~\ref{abla}. We have the following interesting findings. Overall, aligning the vision-text modalities at the early stages (\textit{e.g.} $l_s <6$) often produces higher performance, which again verifies our motivation that effective alignment is crucial for multi-label classification tasks. Besides, aligning from $l_s=1$ may not give the best results as expected. This may be due to the fact that vision and text require specific modality information at the very early layers.

\begin{table}[!t]
    \centering
    \scalebox{0.83}{
    \begin{tabular}{c|c|c|c}
        \toprule[1.5pt]
         Datasets & MS-COCO &Pascal VOC 2007 &NUS-WIDE\\
         \hline 
         PatchCT(label) &88.0 &96.7 &68.0\\
         PatchCT(w/o CT) &84.7 &94.9 &65.8 \\
         \hline
         PatchCT &88.3 &97.1 &68.1\\
         \bottomrule
    \end{tabular}}
    \caption{\small{Ablation results (mAP) of prompt learning and CT alignment. PatchCT(label) denotes the variant that extracts label embedding only using label name, and Patchct (w/o CT) denotes the variant that learns parameters without the CT regularization.}}
    \label{rebuttal}
\end{table}

\paragraph{Effects of $\text{TopK}(\cdot, k)$ Length.}
In our previous experiments, we fix the hyperparameter  $k=200$ in Eq.~\ref{theta} for all images and find it works well in most cases. To further explore whether our model is sensitive to $k$, we report the ablation results with various $k$ on Pascal VOC 2007 at Table.~\ref{kkkk}. We observe that PatchCT is robust to $k$, and one can obtain even better results after finetuning $k$.

\begin{table}[!t]
    \centering
    \begin{tabular}{c|c|c|c|c|c}
        \toprule[1.5pt]
         $k$~ &50 &100 &150 &200 &250\\
         \hline 
         mAP &96.8 &97.2 &96.9 &97.1 &97.0\\
         \bottomrule
    \end{tabular}
    \caption{\small{Ablation on $k$ length in $\text{TopK}(\cdot, k)$ on Pascal VOC 2007 dataset.}}
    \label{kkkk}
\end{table}

\paragraph{Effects of the Prompt Learning.}
In general, prompt learning can help bridge the distribution gap between training and inference. Empirically, CLIP finds that the simple prompt "\textit{A photo of \{label\} .}" often improves performance over the baseline of using only the label text (improves accuracy on ImageNet by 1.3\%). To evaluate the efficiency of prompt learning for multi-label classification, we here report the results of the variant that extracts the textual features only using the label name at Table.~\ref{rebuttal}. Compared to utilizing contextless class names,  prompt learning boost multi-label classification performance on three datasets.

\subsection{Qualitative Analysis} \label{vis}
Another main property of PatchCT is the interpretability brought by the learned transport probabilities, which provide a visualization tool to understand the interactions between labels and patches. Recall that $m$-th column of backward transport plan $\overleftarrow{\Tmat}$ in Eq.~\ref{ct} means the transport probabilities from the $m$-th label to a set of patches. For a backward transport plan of interest $\overleftarrow{\Tmat}_m \in \mathbb{R}^N$, we first normalize it at the range of [0,1], and then reshape it to the grid matrix $\overleftarrow{\Tmat}_m \in \mathbb{R}^{g \times g}$, where $g = \sqrt{N}$ is the number of grids. To visualize the transport plan, we resize the grid matrix to the same size as the raw image via the bicubic interpolation and highlight the most related patches according to the transport probabilities. 
The results are shown at Fig.~\ref{atten} in the following cases: the same image with different labels (the top row), and the same query label with different images (the bottom two rows). From the top row, we find that for a given image, the label queries successfully retrieve the patches that cover the corresponding objects and the transport plan can precisely highlight the main regions of that object. We are also interested in the transport plans in different images. From the middle row, we can see that the transport plans have the ability to adjust the most related regions according to contextual images dynamically. For example, for a given label \textit{zebra}, it tends to focus on the stripes in a close-up image and more on the face in the large background image. We also visualize the top-3 related patches for the \textit{person} query at the bottom row and find that the patches tend to capture the multiple regions of ``person'' in the image. This improves the flexibility and robustness of PatchCT and widens its applications.

\section{Conclusion}
\vspace{-2mm}
In this paper, we reformulate the multi-label image classification as a conditional transport problem and introduce PatchCT for aligning the vision and text modalities under the CT framework, where each image is represented as two discrete distributions over the patches and labels respectively. PatchCT is optimized by minimizing the combined loss that consists of the widely used asymmetric loss and the layer-wise CT distance end-to-end. Extensive experiments on three datasets consistently demonstrate the superiority of the proposed PatchCT. Two ablation studies and visualizations confirm our motivation and the core role of the introduced CT. Since its natural flexibility and simplicity, we hope PatchCT provides innovative ideas for follow-up studies. 

\section{Acknowledgements}
\vspace{-2mm}
This work was supported in part by the National Natural Science Foundation of China under Grant U21B2006; in part by Shaanxi Youth Innovation Team Project; in part by the Fundamental Research Funds for the Central Universities QTZX23037 and QTZX22160; in part by the 111 Project under Grant B18039. 

% \clearpage 

{\small
\bibliographystyle{ieee_fullname}
\bibliography{egbib}
}

\end{document}